\documentclass[runningheads]{llncs}
\usepackage{graphicx}

\usepackage{algorithm}
\usepackage{algorithmic}

\usepackage{graphicx}
\usepackage{amsfonts}

\usepackage{xcolor}

\begin{document}
%
\title{Geolocation of Cultural Heritage using Multi-View Knowledge Graph Embedding}
%
%
\author{Hebatallah A. Mohamed\inst{1} \and 
Sebastiano Vascon\inst{1,2} \and Feliks Hibraj\inst{2} \and 
Stuart James\inst{3} \and Diego Pilutti\inst{1} \and Alessio Del Bue\inst{3} \and 
Marcello Pelillo\inst{1,2} }
%

%
\institute{European Center for Living Technology (ECLT), Ca’ Foscari University of Venice, Italy \and Department of Environmental Sciences, Informatics and Statistics (DAIS), Ca' Foscari University of Venice, Italy \and
Pattern Analysis and Computer Vision (PAVIS), Istituto Italiano di Tecnologia, Genova, Italy}
%
\maketitle              
\begin{abstract}
Knowledge Graphs (KGs) have proven to be a reliable way of structuring data. They can provide a rich source of contextual information about cultural heritage collections. However, cultural heritage KGs are far from being complete. They are often missing important attributes such as geographical location, especially for sculptures and mobile or indoor entities such as paintings. In this
paper, we first present a framework for ingesting knowledge about tangible cultural heritage entities from various data sources and their connected multi-hop knowledge into a geolocalized KG. Secondly, we propose a multi-view learning model for estimating the relative distance between a given pair of cultural heritage entities, based on the geographical as well as the knowledge connections of the entities.




\end{abstract}
\section{Introduction}
The term cultural heritage includes tangible heritage, which can be further specified in \emph{i)} movable (such as paintings, sculptures, coins); \emph{ii)} immovable (such as monuments, archaeological sites), and intangible heritage, such as traditions and performing arts~\cite{maietti2018enhancing}. Preparing cultural heritage collections for exploration by a wide range of users with different backgrounds requires to integrate heterogeneous data into modern information systems. 

Knowledge graphs (KGs) have proven to be a reliable way of structuring data in wide range of domains, including the cultural heritage domain~\cite{Pellegrino2020MoveCH,qassimi2021towards}.
KGs, such as Wikidata~\cite{vrandevcic2014wikidata}, are large directed network of real-world entities and relationships between them, where facts are represented as triplets in the form of (\emph{head entity}, \emph{relation}, \emph{tail entity}). They enable to connect knowledge about cultural heritage collections, and enrich this knowledge with external data coming from heterogeneous sources.

A key challenge for KGs is the human annotation required, resulting in them remaining incomplete. Many general purpose KGs, such as Wikidata~\cite{vrandevcic2014wikidata}, are missing latitude and longitude coordinates for the entities representing tangible cultural heritage (especially the movable ones).
In this paper, we present a method for estimating the relative distance between a given pair of cultural heritage entities in a KG. In particular, we propose a joint learning model to learn comprehensive and representative entity embeddings, based on entity geographical location and semantics. Moreover, we present a framework for ingesting city cultural heritage from multiple general purpose data sources into a geolocalized KG.

In brief, our main contributions are as follows:
\begin{itemize}
  \item We present an ingestion tool and related system framework to create a geolocalized KG for cultural heritage exploration, by ingesting data from heterogeneous data sources based on city coordinates. The tool is based on Neo4j\footnote{\url{https://neo4j.com/}} graph-database and is published as open source on GitHub\footnote{\url{https://github.com/MEMEXProject/MEMEX-KG}}. 
  
  \item We introduce new KG datasets for the geolocation prediction task, with a KG containing both spatial and non-spatial entities and relations.
  
  \item We propose a method that introduces a geospatial distance restriction to refine the embedding representations of geographic entities in a geolocalized KG, which fuses geospatial information and semantic information into a low-dimensional vector space. We then utilize this method for estimating the geographical distance between cultural heritage entities, where we outperform state-of-the-art methods.
\end{itemize}


 



\section{Related Work}
In this section, we discuss work related to cultural heritage KGs, and geolocation using KGs.

\subsection{KGs for Cultural Heritage}
There are some works that have proposed to use KGs to support cultural heritage exploration. For example, Pellegrino et. al. have proposed ~\cite{Pellegrino2020MoveCH},  a general-purpose approach to perform Question-Answering on KGs for querying data concerning cultural heritage. They have assessed the system performance on domain-independent KGs such as Wikidata~\cite{vrandevcic2014wikidata} and DBpedia~\cite{auer2007dbpedia}. The authors of~\cite{qassimi2021towards} have proposed a semantic graph-based recommender system of cultural heritage places, by modeling a graph representing the semantic relatedness of cultural heritage places using tags extracted from the descriptive content of the places from Wikipedia\footnote{\url{https://www.wikipedia.org/}}. 

In addition, there are some existing cultural heritage specific KGs, related to particular places or historical collections. For example, ArCO~\cite{carriero2019arco} is a KG that provides ontology to link people, events, and places about Italian artifacts and document collections. ArCO has been created from different data sources, including general purpose KGs such as Wikidata and DBpedia. Linked Stage Graph~\cite{tietz2019linked} which organizes and interconnects data about the Stuttgart State Theaters. ArDO~\cite{vsesviatska2021ardo} is an ontology to represent the dynamics of annotations of general archival resources. 
In our work, we propose a system framework for ingesting knowledge about any city cultural heritage using heterogeneous data coming from Wikidata~\cite{vrandevcic2014wikidata}, and any other geolocalized linked open data (LOD) such as Europeana~\cite{freire2019aggregation}, into a geolocalized KG.

\subsection{Geolocation using KGs}
There have been only few works that have focused on the task of geographical KG construction. The Guo et. al. have proposed GeoKG~\cite{guo2021method}, a system framework for extracting geographical knowledge from geographic datasets and represent the knowledge by means of concepts and relations with the aid of GeoSPARQL\footnote{\url{http://www.opengis.net/ont/geosparql}}. However, this work has focused on specific semantic relationships such as \emph{subclass\_of} or \emph{instance\_of} to describe the subclass concepts of the geographical entities, such as \emph{river} and \emph{railway station}. In our work, we aim at constructing a geolocalized KG where tangible cultural heritage entities can be represented with richer semantic relations, such as \emph{architectural\_style}, \emph{architect} and \emph{exhibited\_at}. 


Moreover, there are works that focused on geographical KG completion and location prediction. For example, the authors of~\cite{qiu2019knowledge} have proposed a translational embedding model that has the ability to capture and embed the geospatial distance restriction with the semantic information of the geographical KG into a vector space. Then, the optimized model outputs the refined representations of geo-entities and geo-relations, which improves the completion performance on
the sparse geographical KG. However, this work does not consider the existence of non geo-entity and non geo-relations in the KG. In our work, we aim at predicting the distance between two entities in a KG that contains both geographical and non-geographical entities and relations.



\section{Data Curation for Geolocalized Knowledge Graph} \label{sec:ingestion}
In this section, we discuss details about our developed ingestion tool, and the related system framework for ingesting data about a given city and the cultural heritage it contains into a geolocalized KG.

\subsection{Data Sources}
In this work, Wikidata\footnote{\url{https://www.wikidata.org/}} and Europeana\footnote{\url{https://www.europeana.eu/}} have been selected for enriching our geolocalized KG, however other general purpose KGs and LOD can be incorporated. 
Wikidata is a public crowd-sourced knowledge base, with facts and figures about tens of millions of items (93,207,191). The data are offered freely, with no restriction on the reuse and modification (Creative Commons Zero). 
Wikidata provides its data openly in an easy to use standard Resource Description Framework (RDF). It provides much of its data in hundreds of languages. Already, large amounts of data about cultural assets are being shared with Wikidata by formal partnerships with Galleries, Libraries, Archives and Museums (GLAMS).  

On the other hand, Europeana aims to facilitate the usage of digitized cultural heritage resources from and about Europe~\cite{haslhofer2011data}. It seeks to enable users to access content in all European languages via the Europeana collections portal and allow applications to use cultural heritage metadata via its open APIs. Europeana holds metadata from over 3,700 providers~\cite{freire2019aggregation}, mostly GLAMS. However, it uses a strictly federated ontology limiting the diversity of the meta-data.

\begin{figure}[t!bp]
\begin{center}
\includegraphics[width=0.95\textwidth]{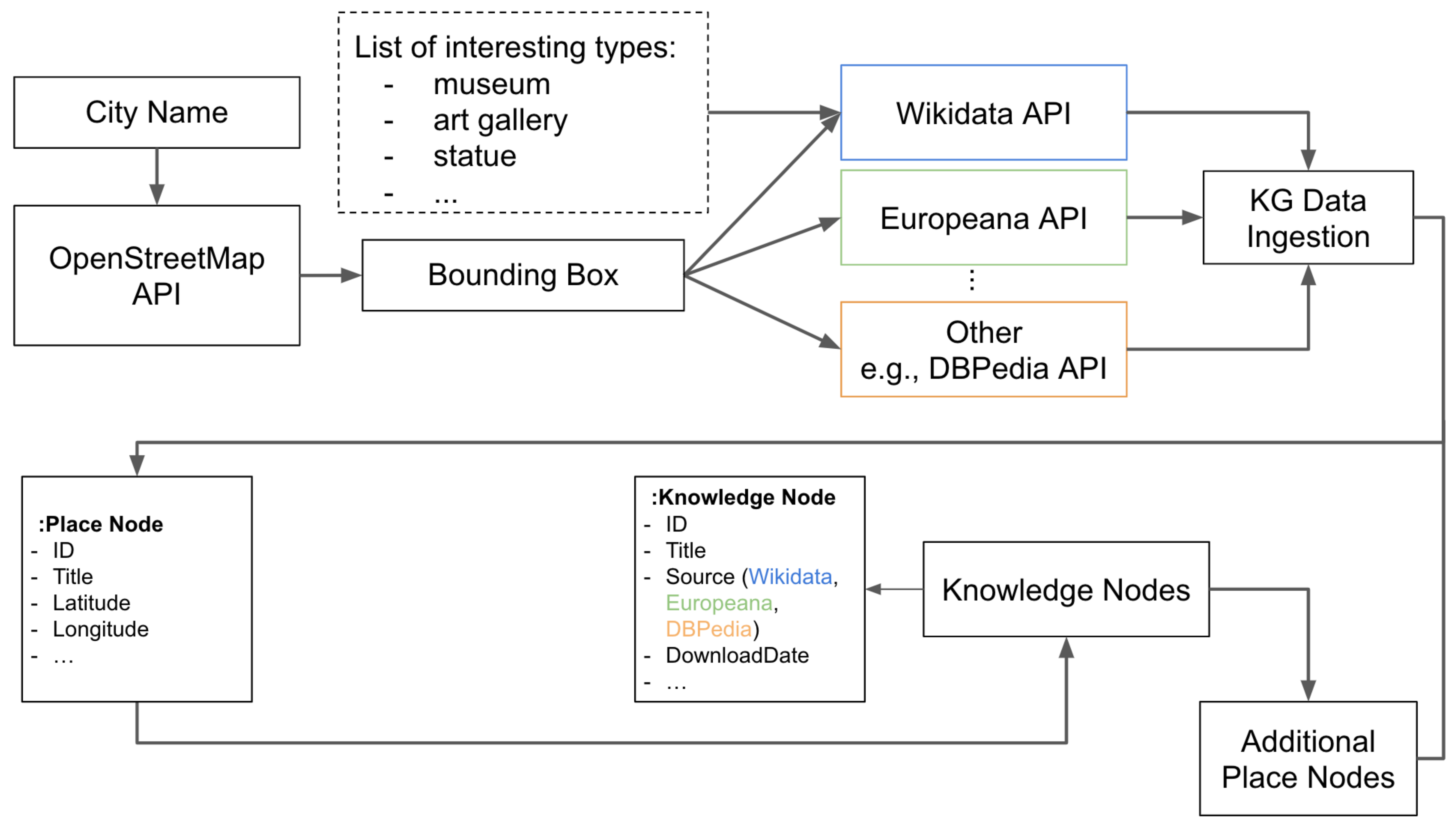}
\caption{Information extraction workflow.}
\label{fig:scheme}
\end{center}
\vspace{-0.5cm}
\end{figure}

\subsection{Information Extraction Workflow}
We propose a localized approach that grows a KG based on association to known landmarks.
We define two types of KG nodes: \emph{Place} and \emph{Knowledge}. The former represents tangible data while the latter represents intangible data. 
As shown in Figure \ref{fig:scheme}, first, we use OpenStreetMap (OSM)\footnote{\url{https://www.openstreetmap.org/}} to get the bounding box of the target city. Then coordinates are used to query Wikidata to retrieve geo-entities within the selected scope, using Wikidata \emph{coordinate\_location (P625)} attribute. Moreover, a list of interesting entity types can be predefined by the user, to allow the ingestion tool to focus on specific cultural heritage related entities. We use the Wikidata \emph{instance\_of (P31) and subclass\_of (P279)} attributes to select entities with specific types.
These form the basis of our $Places$ (node type is $Place$) where related additional properties are downloaded. 

\begin{figure}[t!bp]
\begin{center}
\includegraphics[width=0.75\textwidth]{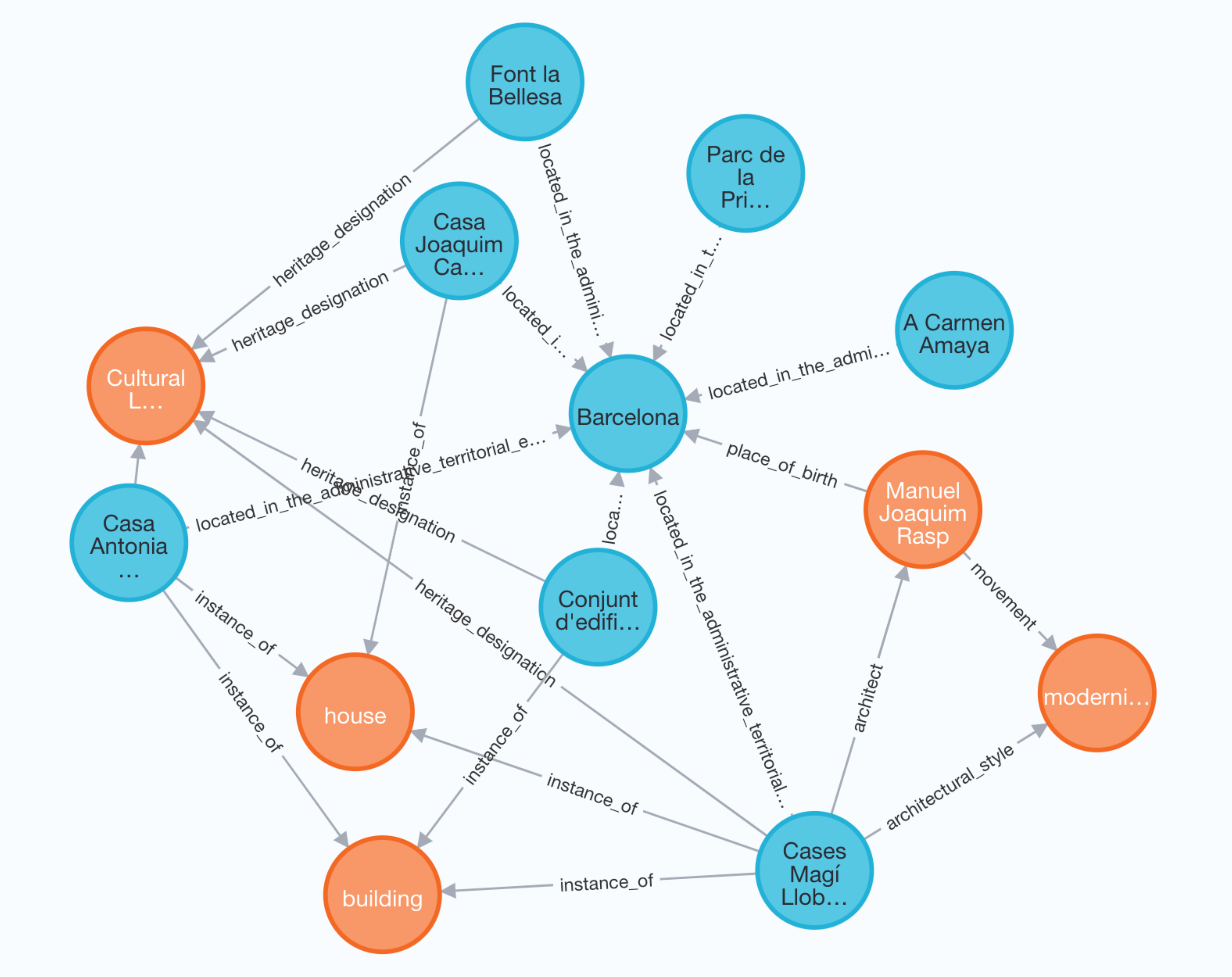}
\caption{A subset of our geolocalized KG (orange nodes are $Knowledge$ and cyan ones are $Places$).}
\label{fig:geokg}
\end{center}
\vspace{-0.5cm}
\end{figure}

From each of the identified $Places$, we then query associated nodes based on all relations from Wikidata, this new set of nodes we refer to as $Knowledge$. This could be for example, the architect of a building. We repeat this step searching for new relationships a predefined number of times referred to as hops. We use the same approach to ingest data from Europeana. For a given city, we retrieve its coordinates using OSM, then a query is performed against Europeana API to identify content within the GPS region to add nodes in the KG. Figure \ref{fig:geokg} shows a small subset of our geolocalized KG.

\subsection{NER for Data Integration}
We utilize Named Entity Recognition (NER) technique in order to link KG entities from different sources of information.  
More precisely, we use spaCy\footnote{\url{https://www.spacy.io/}}, a Python library with wide linguistic features. SpaCy annotates text with different types of named entities. We focus on the  types listed in Table \ref{tab:ner}.
For each entity ingested from Europeana, we apply NER to extract named entities from the title of Europeana entity. Then, we match those extracted named entities with the titles of Wikidata entities.
If there is a match found, we create a relation (we name it $related\_to$) between the matching Europeana and Wikidata entities. For example, when applying NER on a Europeana entity titled ``\emph{Carmen Amaya’s last dance}", the named entity ``\emph{Carmen Amaya}" of type PER will be extracted, and the Europeana entity will be linked to the entity titled ``\emph{A Carmen Amaya}" from Wikidata in the KG.

\begin{table}[t!bp]
  \caption{Named entity types used to integrate data from different sources.}
  \label{tab:ner}
  \centering
  \begin{tabular}{l|l}
    \hline
    Type & Meaning \\
    \hline
    GPE & Geopolitical entity, i.e., countries, cities, states. \\
    LOC & Non-GPE locations, mountain ranges, bodies of water. \\
    FAC & Buildings, airports, highways, bridges, etc. \\
    PER & People, including fictional ones. \\
    \hline
  \end{tabular}
\end{table}






\section{Geolocation using Multi-view Graph Embedding} 
\begin{figure}[t!bp]
\begin{center}
\includegraphics[width=0.99\textwidth]{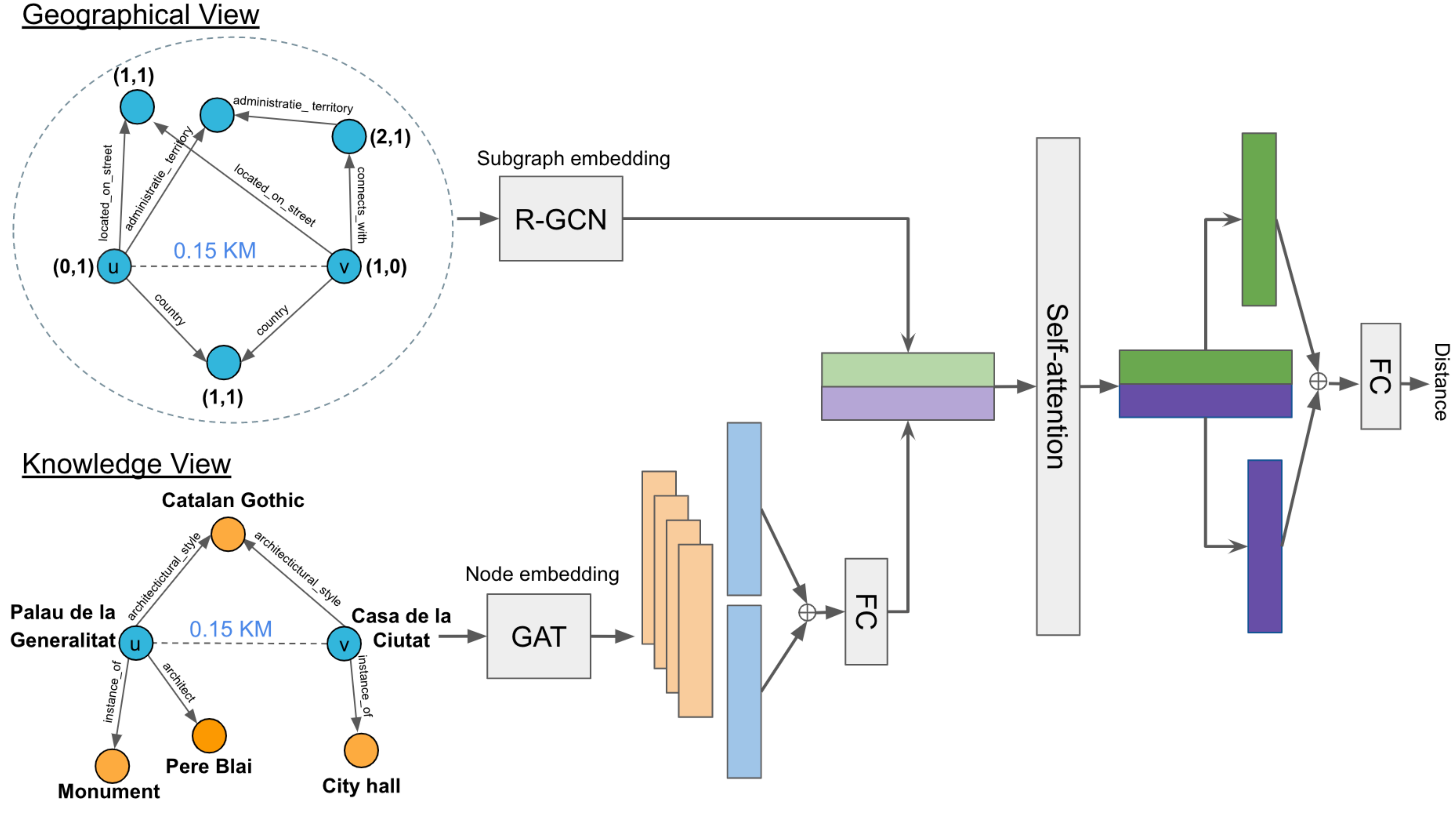}
\caption{The proposed approach for geographical location prediction.}
\label{fig:proposed}
\end{center}
\vspace{-0.5cm}
\end{figure}

Given a geolocalized KG $G$, composed of a large number of triples in the form of (\emph{head entity}, \emph{relation}, \emph{tail entity}). All entities have associated type $\in$ \{$Place$, $Knowledge$\}, title $S = [w_1, w_2, ..., w_n]$ with $n$ words. Additionally, most of the $Place$ entities have a latitude and longitude information. From our KG we aim to predict the distance $d_t$ between any two $Place$ entities, $u$ and $v$, where the latitude and longitude property is missing in at least one of those entities (i.e., the distance between is unknown).
To achieve this, we propose a method to construct two types of $Place$ entity correlations: 1) based on the geographical view using the $Place$ nodes, and 2) based on entity semantics using the $Knowledge$ nodes. An overview about our proposed approach is shown in Figure \ref{fig:proposed}, and more details are described in the following sections.

\subsection{Geographical View}
\label{sec:geo_view}

The geographical view aims at capturing the geographical relation between two geo-entities, $Places$, in the KG. To represent this view, we extract an induced subgraph around the two target entities. The subgraph represent how the two geo-entities are connected to each other, in terms of direct links or common nodes between the two target nodes. Since we are focusing on geographical view, we restrict the common nodes to \emph{Place} node type. We then embed the extracted subgraph with a relational graph convolution network (R-GCN)~\cite{schlichtkrull2018modeling} to represent the geographical relation between the target entities. In more details, we perform the following steps 1. Subgraph Extraction (sec.~\ref{sec:subgraph_extract}), and 2. Node Representation (sec.~\ref{sec:node_representation}).

\subsubsection{Subgraph Extraction}\label{sec:subgraph_extract}
We assume that the paths (i.e., geo-entities and geo-relations) on the KG connecting the two target nodes contain the information that could imply the distance between the two target entities. Hence, as a first step, we extract an enclosing subgraph around the target nodes. The enclosing subgraph between nodes $u$ and $v$
is the graph induced by all the nodes that occur on a path between $u$ and $v$. The subgraph set is given by the intersection of neighbors
of the two target nodes followed by a pruning procedure. 

More precisely, let $N_k(u)$ and $N_k(v)$ be set of nodes in the
$k$-hop (undirected) neighborhood of the two target nodes in
the KG. We compute the enclosing subgraph by taking the
intersection, $N_k(u) \cap N_k(v)$, of these $k$-hop neighborhood
sets and then prune nodes that are isolated or at a distance
greater than $k$ from either of the target nodes. This results in all the nodes that occur
on a path of length at most $k + 1$ between nodes $u$ and $v$, where we refer to it as the induced subgraph $G(u, v)$.

\subsubsection{Node Representation} \label{sec:node_representation}
We define an embedding for each entity (node) in the subgraph $G(u, v)$. Following~\cite{Zhang2018LinkPB}, each node $i$ in the subgraph is labeled with the tuple $(d(i, u), d(i, v))$, where $d(i, u)$ is the shortest distance between nodes $i$ and $u$ (likewise for $d(i, v)$). The two target nodes, $u$ and $v$, are labeled (0, 1), and (1, 0) to be identifiable by the model. This scheme captures the position of each node in the subgraph with respect to the target nodes, as shown in Figure \ref{fig:proposed}. The node features are defined as [one-hot$(d(i, u)) \oplus $ one-hot$(d(i, v))]$, representing the concatenation of the one-hot embedding of the labels.

\subsubsection{Geographical View Embedding}
We use multiple layers of the multi-relational R-GCN~\cite{schlichtkrull2018modeling} to learn the embeddings of the extracted subgraph $G(u, v)$. R-GCN adopts general message-passing scheme~\cite{Xu2019HowPA}, where a node representation is iteratively updated by combining it with the aggregation of its neighbors' representation. The subgraph representation of $G(u, v)$ is obtained by average-pooling of all the node representations:
In the $k$-th layer of our graph neural network (GNN), $a_i^{k}$ represents the aggregated message from the neighbors of node $i$.
The aggregation function is defined as:
\begin{equation}
\textstyle 
a_i^{k}=\sum_{r=1}^{R}\sum_{s \in N_r(i)}\alpha_{r(s,i)}^{k}W_r^{k}h_s^{k-1},
\end{equation}
where $R$ is the total number of unique relations, $N_r(i)$ represents the neighbors of node $i$ under relation $r$, $W_r^{k}$ is the transformation matrix of the $k$-th layer over relation $r$, and $\alpha_{r(s,i)}$ is the edge attention weight at the $k$-th layer corresponding to the edge between nodes $s$ and $i$ via relation $r$.

The latent representation of node $i$ in the $k$-th layer is:
\begin{equation}
\textstyle 
h_i^{k}=ReLU (W_0^{k}h_i^{k-1} + a_i^{k}),
\end{equation}
where $W_0$ aims at retaining the information of the node itself using self-connection, and $ReLU$ is an activation function.

The subgraph representation of $G(u, v)$ is obtained by average-pooling of all the node representations:
\begin{equation}
\textstyle 
h^{L}_{G(u, v)} = \frac{1}{|V|}\sum_{i \in V}h_i^{L},
\end{equation}
where $V$ denotes the set of nodes in $G(u, v)$, and $L$ represents the number of layers of message-passing.

\subsection{Knowledge View}
The knowledge view aims at representing the semantic relation between two target geo-entities, $Places$, in the KG, by means of their surrounding $Knowledge$ nodes. We employ a graph attention network (GAT)~\cite{velivckovic2017graph} for representing each target entity semantics based on entity textual title. Then, we concatenate the representation of the two target entities and pass the concatenated representation into a linear layer. In more details, as with Geographical View, we perform: 1. Node Representation (sec.~\ref{sec:knowledge_node_representation}), and 2. Node Embedding (sec.~\ref{sec:knowledge_node_embedding}).

\subsubsection{Node Representation}\label{sec:knowledge_node_representation}
For each node in $G$ with title $S = [w_1, w_2, ..., w_n]$ ($n$ is different for each node since each title might have different length), we initialize the embedding of each word $w$ with a pre-trained word vector from GloVe~\cite{pennington2014glove}. Let $p$ denote the dimension of each word vector. We obtain the sentence representation $\overrightarrow{S}$ by aggregating the embedding of each word, where $\overrightarrow{S} \in \mathbb{R}^{p}$. In aggregation, we use only simple averaging due to its validity~\cite{zhu2018exploring}.

\subsubsection{Node Embedding}\label{sec:knowledge_node_embedding}
We employ a GAT to learn representations of nodes. GAT applies attention mechanism on graph-structured data. It updates the representation of a vertex by propagating information to its neighbors, where the weights of its neighbor nodes is learned by attention mechanism automatically. 
Formally, given the input vertex features $\left\{\overrightarrow{S_1}, \overrightarrow{S_2}, \ldots, \overrightarrow{S_n}\right\}$, a GAT layer updates the
vertex representations by following steps:

\begin{equation}
e_{i j}=\exp \left(ReLU\left(a^{T}\left[W_{k} \overrightarrow{S_i} \oplus W_{k} \overrightarrow{S_j}\right]\right)\right)
\end{equation}

\begin{equation}
\alpha_{i j}=softmax_{j}\left(e_{i j}\right)=\frac{\exp \left(e_{i j}\right)}{\sum_{k \in N_{i}} \exp \left(e_{i k}\right)}
\end{equation}

\begin{equation}
h_{i}^{\prime}=\sigma\left(\sum_{j \in N_{i}} \alpha_{i j} W_k h_{j}\right)
\end{equation}
where $W_k$ and $a$ are learnable parameters, $\oplus$ is the concatenation operation. 

\subsubsection{Knowledge View Embedding}
After calculating the embedding of each node in the graph, we concatenate the embeddings of the target entities, $u$ and $v$. Then, we pass these concatenated representations through a linear layer, given by
\begin{equation}
\textstyle 
{h}_{u,v}=[{h}_{u}^{\prime} \oplus {h}_{v}^{\prime}]W_s,
\end{equation}
where $\oplus$ refers to the concatenation operation, and $W_s$ is a learnable weight parameter.

\subsection{Multi-View Embedding}
In order to enable the cooperation among the different views during the learning process and effectively fuse multi-view representations, our model enables information sharing across all views via a scaled dot-product self-attention~\cite{vaswani2017attention}. 
Given the representations from the different views as $\left\{\mathcal{E}_{i} \right\}_{i=1}^{2}$, we stack them into a matrix $X \in \mathbb{R}^{2 \times d}$ of dimension $d$. We associate query, key and value matrices $Q$, $K$ and $V$ as follows:

\begin{equation}
\quad Q=X W_{Q}, \quad K=X W_{K}, \quad V=X W_{V}
\end{equation}
where $W_{Q}$, $W_{K}$ and $W_{V}$ are weight matrices. We then propagate information among all views as follows:

\begin{equation}
Y = softmax\left(\frac{Q K^{T}}{\sqrt{d_k}}\right) V
\end{equation}
$\hat{\mathcal{E}}_{i}$ will then be the $i$-th row in matrix $Y$, which considered as the relevant global information for $i$-th view. To incorporate this information in the learning process, we concatenate the representations of the different views (geographical and knowledge views) and pass it to a fully connected (FC) layer, to predict the distance $\hat{d_t}$ as follows:
\begin{equation}
\textstyle 
\hat{d_t}=[\hat{\mathcal{E}}_{1} \oplus \hat{\mathcal{E}}_{2}]W_m,
\end{equation}
where $\oplus$ refers to the concatenation operation, and $W_m$ is a learnable weight parameter.





\subsection{Learning Objective}
Given the ellipsoidal shape of the earth’s surface, we apply the Haversine distance~\cite{robusto1957cosine} to calculate the distance of two points represented by their latitude in range of $\{-90,90\}$ and longitude in range of $\{-180,180\}$. The Haversine distance is the great circle distance between two geographical coordinate pairs.
We train our model to reduce the Mean Squared Error (MSE) loss based on the actual and predicted Haversine distance, $d_t$ and $\hat{d_t}$. 

\begin{figure}[t!bp]
\begin{center}
\includegraphics[width=0.99\textwidth]{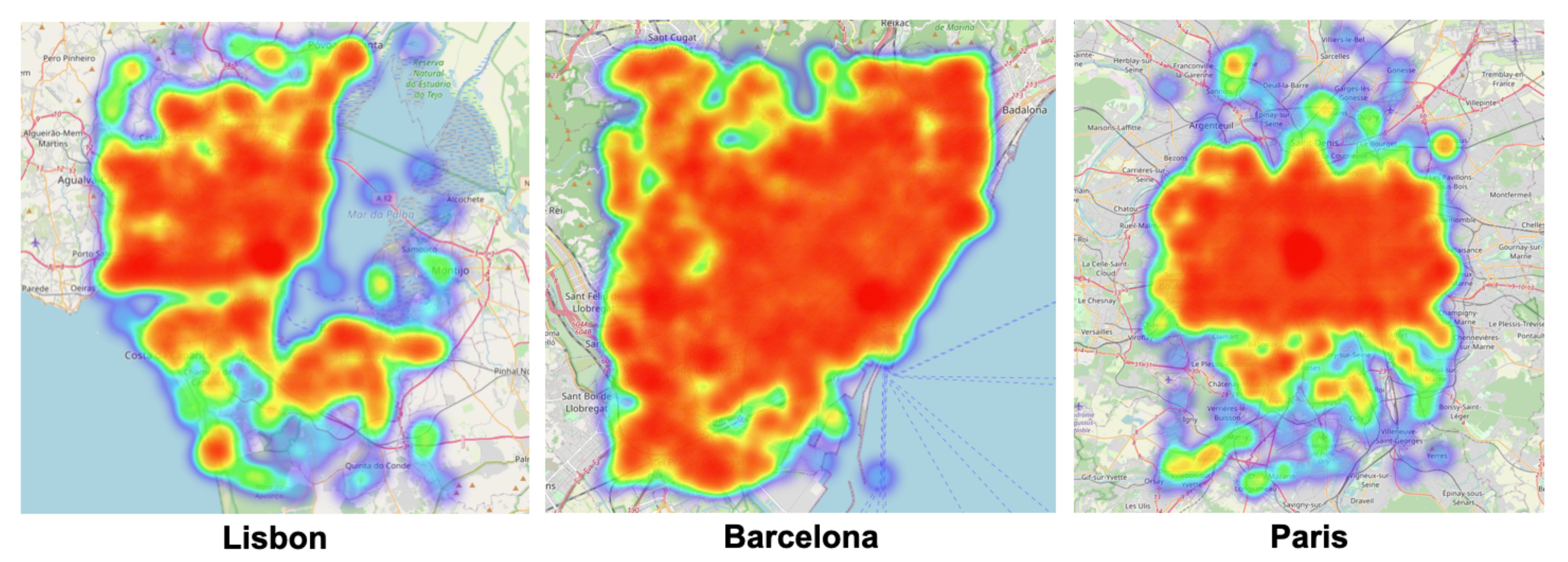}
\caption{Visualization of tangible cultural heritage on the map.}
\label{fig:datasets}
\end{center}
\vspace{-0.5cm}
\end{figure}

\section{Experiments}

\subsection{Datasets}
In order to evaluate our model, we create three datasets using our ingestion tool described in section \ref{sec:ingestion}. The datasets are about tangible ($Place$) and $Knowledge$ of three main cities in Europe: Lisbon, Barcelona and Paris. 

We set the number of hops to 3, since 3-hop contains sufficient information about each tangible cultural heritage entity. 
Statistics about the datasets, including the maximum distance between geo-entites representing cultural heritage related places are shown in Table \ref{tab:data}. The maximum distance between the places in Barcelona dataset is less than in Lisbon and Paris. Lisbon dataset is the most challenging since it contains fewer number of nodes and relations. Moreover, Lisbon has the highest maximum distance between places. In Figure \ref{fig:datasets}, we visualise the location of the tangible cultural heritage of the different cites. We randomly split the links between $Places$ into 80\% and 20\% sets with corresponding distances for training and testing, respectively. Hyper-parameters are optimised for the training set.

\begin{table}[t!bp]
  \caption{Statistics of the city datasets.}
  \label{tab:data}
  \centering
  \begin{tabular}{l|c|c|c}
    \hline
    & Lisbon & Barcelona & Paris \\
    \hline
    \#$Place$ Nodes & 2,780 & 5,989 & 26,214 \\
    \#$Knowledge$ Nodes & 5,860 & 9,140 & 35,885 \\
    \#Links & 21,253 & 37,384 & 196,898 \\
    \#Relation types & 377 & 708 & 578 \\
    \hline
    \#$Place$-$Place$ links & 5,077 & 6,907 & 48,954 \\
    Maximum distance & 35.32 KMs & 16.76 KMs & 34.69 KMs  \\
    \hline
  \end{tabular}
\end{table}


\subsection{Baselines and Implementation Details}
We compare our model against GAT~\cite{velivckovic2017graph} and R-GCN~\cite{schlichtkrull2018modeling} as baselines when textual title of the $Place$ and $Knowledge$ nodes are used for message-passing to generate node embeddings. We also evaluate our model when using only the geographical view (Ours - geographical), utilizing enclosing subgraph embeddings with R-GCN. Moreover, we experiment our model without the attention layer (Ours - without att). 

The latent embedding sizes used in all our models are set to 32. We set the number of layers in GAT and R-GCN to 3. To train our model, we use Adam optimizer~\cite{kingma2014adam} with a learning rate of 1e-4. We run our experiments on a machine with two Intel Xeon Gold 6230 CPUs running at 2.10 GHz with 128 GB of memory, and Nvidia Quadro RTX 5000 GPU with 16 GB of memory.
Finally, we use Mean Absolute Error (MAE) and Root Mean Square Error (RMSE) to measure the prediction errors.

\subsection{Results}
In Table \ref{tab:results}, we present the results of the different prediction methods in terms of MAE and RMSE. As shown, our proposed method outperform other baseline methods. Best results are achieved on Barcelona dataset, since the maximum distance between the places is less than the one in Lisbon and Paris.

The results show that GAT is outperforming R-GCN when applying the message-passing technique on the whole KG for extracting node embeddings. This indicates that GAT is able to focus on certain neighbors to represent a $Place$ node due to its attention mechanism. However, when applying R-GCN on enclosing subgraphs (Ours - geographical), the performance is outperforming the R-GCN baseline since the model is focusing on fewer unique relation types representing the spatial relation. 
Finally, the results show that applying an attention layer helps to fuse the multiple views in our proposed model.

\begin{table}[t!bp]
  \caption{The results showing MAE and RMSE of Haversine distance in $km$. Best results are in \textbf{Bold}.}
  \label{tab:results}
  \centering
  \begin{tabular}{l|c|c|c|c|c|c}
    \hline
     &\multicolumn{2}{c}{Lisbon} & \multicolumn{2}{c}{Barcelona} & \multicolumn{2}{c}{Paris} \\
    \hline
     & MAE & RMSE & MAE & RMSE & MAE & RMSE\\
    \hline
    GAT & 1.95 & 2.64 & 1.43 & 1.87 & 0.70 & 1.23 \\
    R-GCN  & 2.30 & 3.12 & 1.48 & 1.90 & 1.21 & 1.63\\
    \hline
    Ours - geographical & 2.23 & 3.10 & 0.48 & 0.75 & 0.83 & 1.42 \\
    Ours - without att. & 2.13 & 2.86 & 0.56 & 0.86 & 0.67 & 1.20 \\
    \hline
    Ours & \textbf{1.90} & \textbf{2.50} &  \textbf{0.42} & \textbf{0.71} & \textbf{0.59} & \textbf{1.05} \\
    \hline
  \end{tabular}
\end{table}

\section{Conclusion}
We present an ingestion tool and a framework to create a geolocalized KG for contextualising cultural heritage. In addition, we propose a method that introduces a geospatial distance restriction to refine the embedding representations of geographic entities in a geolocalized KG, which fuses geospatial information and semantic information into a low-dimensional vector space. We utilize this method for a geographical distance prediction task, where we outperform baseline methods. 



\section*{ACKNOWLEDGMENTS}
This work was supported by MEMEX project funded by the European Union’s Horizon 2020 research and innovation program under grant agreement No 870743.

%
%
%
\bibliographystyle{splncs04}
\bibliography{mybibliography}

\end{document}